\documentclass[conference]{IEEEtran}
\IEEEoverridecommandlockouts
\usepackage{cite}
\usepackage{amsmath,amssymb,amsfonts}
\usepackage{algorithmic}
\usepackage{graphicx}
\usepackage{textcomp}
\usepackage{xcolor}
\usepackage{caption}
\usepackage{subcaption}
\usepackage{booktabs}
\usepackage{multirow}
\usepackage{multicol}
\def\BibTeX{{\rm B\kern-.05em{\sc i\kern-.025em b}\kern-.08em
    T\kern-.1667em\lower.7ex\hbox{E}\kern-.125emX}}
\begin{document}

\title{STING: Self-attention based Time-series Imputation Networks using GAN}

\author{\IEEEauthorblockN{Eunkyu Oh, Taehun Kim, Yunhu Ji, Sushil Khyalia}
\IEEEauthorblockA{\textit{Samsung Research, Samsung Electronics} \\
Seoul, Republic of Korea \\
\{eunkyu1.oh, taehun33.kim, yunhu.ji, sus.hil\}@samsung.com}
}

\maketitle

\insert\footins{\noindent\footnotesize \copyright\, 2021 IEEE. Personal use of this material is permitted. Permission from IEEE must be obtained for all other uses, in any current or future media, including reprinting/republishing this material for advertising or promotional purposes, creating new collective works, for resale or redistribution to servers or lists, or reuse of any copyrighted component of this work in other works.}

\begin{abstract}
Time series data are ubiquitous in real-world applications. However, one of the most common problems is that the time series data could have missing values by the inherent nature of the data collection process. So imputing missing values from multivariate (correlated) time series data is imperative to improve a prediction performance while making an accurate data-driven decision. Conventional works for imputation simply delete missing values or fill them based on mean/zero. Although recent works based on deep neural networks have shown remarkable results, they still have a limitation to capture the complex generation process of the multivariate time series. In this paper, we propose a novel imputation method for multivariate time series data, called STING (Self-attention based Time-series Imputation Networks using GAN). We take advantage of generative adversarial networks and bidirectional recurrent neural networks to learn latent representations of the time series. In addition, we introduce a novel attention mechanism to capture the weighted correlations of the whole sequence and avoid potential bias brought by unrelated ones. Experimental results on three real-world datasets demonstrate that STING outperforms the existing state-of-the-art methods in terms of imputation accuracy as well as downstream tasks with the imputed values therein.
\end{abstract}


\section{Introduction}
Multivariate time series data are everywhere and continuously generated every day. Many real-time application domains have analyzed these signals for predictive analytics. For instance, there are financial marketing on forecasting the stock price \cite{hsieh2011forecasting}, predicting medical diagnosis of patients \cite{che2018recurrent, liu2016learning}, weather forecasting \cite{shi2015convolutional, rani2012recent}, and real-time traffic prediction \cite{wang2017deepsd, zhang2016deep}. However, it is inevitable that the time series data contain missing values for some reasons, such as certain data features being collected later or records being lost due to equipment damage or communication errors. In the medical field, certain information collected e.g., from a biopsy may be difficult or even dangerous to obtain \cite{yoon2018personalized}. These kinds of missing data significantly degrade the model quality and even make wrong judgments by introducing a substantial amount of bias \cite{Lall2016HowMI}. So, imputing missing values in time series has become a paramount issue for making accurate data-driven decisions.

Conventional methods of imputing missing values can be classified into two categories: a discriminative method and a generative method. The former includes Multivariate Imputation by Chained Equations (MICE) \cite{buuren2010mice} and MissForest \cite{stekhoven2012missforest} while the latter includes deep neural network based algorithms (e.g., Denoising Auto Encoders (DAE) and Generative Adversarial Networks (GAN)) \cite{vincent2008extracting, Gondara2017MultipleIU}. However, these methods have been developed for non-time series data so that the temporal dependencies between observations in the time series could be rarely considered. In particular, DAE needs fully complete data for a training phase but this requirement is almost impossible as missing values are part of the inherent structure of the problem. Recent state-of-the-art works for the time series imputation are based on Recurrent Neural Networks (RNN) and GAN \cite{che2018recurrent, luo2018multivariate, cao2018brits, luo2019e2gan}. They capture temporal dependencies with various aspects of the observed (or missing) data characteristics, such as time decay, feature correlation, and temporal belief gates.

Inspired by existing deep generative models showing remarkable advancement, we propose a novel imputation method for multivariate time series data, called STING (Self-attention based Time-series Imputation Network using GAN). We take the generative adversarial networks as the base architecture which could estimate a true data distribution while imputing the original incomplete time series data. To be specific, the generator learns the underlying distribution of multivariate time series data to accurately impute the missing values, and the discriminator learns to distinguish between observed and imputed elements. The generator in GAN internally adopts the novel RNN cell, GRUI (GRU for Imputation) \cite{luo2018multivariate} to learn the latent relationships between observations with non-fixed time lags. It weighs the impact on the past observations according to the time lags. To leverage information from both future and past observations to impute the current missing values, Bidirectional RNN (B-RNN) is adopted to estimate the variables from both forward and backward directions. In addition, we propose a novel attention mechanism to pay selective attention to highly related information in each time series. This allows efficient training when the time series sequence is long and the time interval between two observations is large. Experiment results on three real-world datasets show that STING outperforms state-of-the-art methods in terms of imputation performance. Our model is also superior to the baselines in a post-imputation task as an indirect measure of imputation performance.

\section{Related Work}

Che et al.\cite{che2018recurrent} proposed Gated Recurrent Unit with Decay (GRU-D) that learns missing patterns of incomplete time series and predicts the target labels while imputing missing values in a healthcare dataset. The model takes into account a weighted combination of its last observation and the empirical mean using RNN. They introduced an input decay rate and a hidden state decay rate to control the decay factor. However, this model has a fundamental premise that missing patterns of data are often correlated with the target labels (i.e., informative missingness). By taking this fact into account, they take a unified approach by integrating the imputation and prediction task (i.e., of target labels) into one process. This makes the model less generalized due to the dependency on whether the target labels are completely observed. Therefore, it is hard to be used in unsupervised settings without labels \cite{che2018recurrent} or when the labels are not clear. They also impose the statistical assumption that the imputed values are the ratio of the last observation and the empirical mean.

As a similar kind of research, Cao et al.\cite{cao2018brits} proposed an RNN-based method, called Bidirectional Recurrent Imputation for Time Series (BRITS) to directly learn the missing values considered as latent variables of the bidirectional RNN graph. This combines historical-based estimation and feature-based estimation for feature correlations and applies a learning strategy that makes missing values get delayed gradients. However, it aims to predict target labels based on the given time series while learning imputation simultaneously. So it needs to know target labels in a training phase. However, as the target labels could be unknown or contain missing values, this is a quite strong constraint. In consequence, the imputation performance is highly sensitive to the integrity of the target labels. Unlike \cite{che2018recurrent} and \cite{cao2018brits}, our model is independent of the target labels during any process.

Luo et al.\cite{luo2018multivariate} proposed GAN-based imputation model, called GAN-2-stages. In order to model the distribution with temporal irregularity, Gated Recurrent Unit for Imputation (GRUI) is proposed to learn to decay the influence of the past observations according to how long the time has passed in an irregular time interval. They further train an input "noise" of the generator to find the best noise from the latent input space so that the generated sample becomes the most similar to the original one. However, while the original samples have different time decays by the missingness, the generated ones have the same time decays due to their completeness. This distinct difference in the time decays makes the discriminator easy to distinguish between fake data and real data and prevents the stable training of the generator because of the faster convergence of the discriminator. Moreover, they take a self-feed training method where incorrect outputs during training will continue to be reflected in subsequent learning until the end.

As the follow-up work, Luo et al.\cite{luo2019e2gan} proposed End-to-End Generative Adversarial Network (E$^2$GAN). They exploit a compression and a reconstruction strategy to avoid the "noise" optimization stage by using a denoising auto-encoder \cite{vincent2008extracting}. In the generator, a random noise is added to an original time series as an input and the encoder tries to map the input into a low-dimensional vector. Then, the decoder reconstructs it from the low-dimensional vector to generate a sample. Through this process, E$^2$GAN could force to learn a compressed representation of the input while learning the distribution of the original time series. However, E$^2$GAN still holds the limitations of GAN-2-stages such as the faster convergence of the discriminator compared to the generator and the self-feed training of RNN. Unlike \cite{luo2018multivariate} and \cite{luo2019e2gan}, we take bidirectional delayed gradients \cite{cao2018brits} for quickly and efficiently training RNN models in the generator that use the observed values. Also, stable adversarial learning is possible because the proposed discriminator tries to solve a more specific problem by distinguishing if each element of the input matrix is either true value (observed) or fake value (generated), instead of the whole input matrix itself.

\begin{figure*}[t]
    \centering
    \includegraphics[width=2\columnwidth,keepaspectratio]{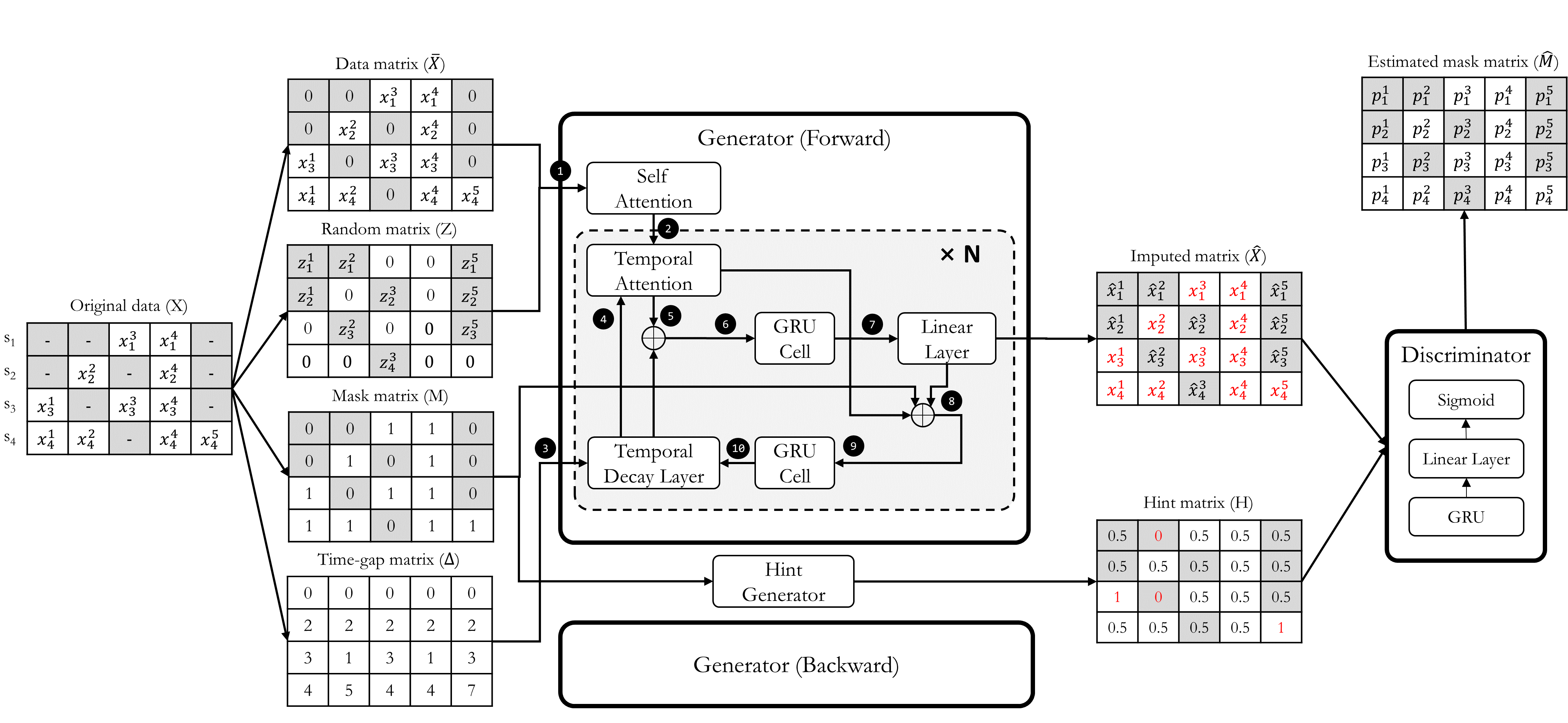}
    \caption{The architecture of STING. It consists of two generators for dealing with a forward/backward direction and one discriminator. All modules basically adopt RNN structure to iteratively process each time step in the time sequence with N length. The gray-colored elements in each matrix correspond to the missing values of the original data. The data flow of the backward generator is omitted as it has the same internal structure as the forward one except for processing the data in the opposite direction.}
    \label{fig:archi}
\end{figure*}

\section{Problem Definition and Notations}
We denote multivariate time series data $X = \left\{x_1, x_2, ..., x_T\right\}$ as a sequence of T observations and the t-th observation $x_t \in \mathbb{R}^D$ consists of D features, $\left\{x_t^1, x_t^2, ..., x_t^D\right\}$. That is, $x_t^d$ is denoted as the value of d-th variable of $x_t$. $X$ is an incomplete matrix with missing values. We also introduce a mask vector $m_t$ to denote positions where variables are missing in $x_t$. So, $m_t^d$ is defined as below.
$$m_t^d = 
\begin{cases}
\; 1 & \mbox{if }x_t^d\mbox{ is observed} \\
\; 0 & \mbox{otherwise}
\end{cases}
$$
We define $\bar{X}$ that is almost the same as $X$ except that if $x_t^d$ is a missing value, $\bar{x}_t^d$ is zero. $z_t$ is a random vector with the same dimension as $x_t$. We denote $s_t$ as a timestamp when the t-th values are observed, which is an element of $S$. Time intervals of timestamps may not be the same, so we define $\delta_t^d$ as the time interval from the last observation to the current timestamp $s_t$.
$$\delta_t^d = 
\begin{cases}
\; s_t - s_{t-1} + \delta_{t-1}^d & \mbox{if }t > 1, m_{t-1}^d = 0 \\
\; s_t - s_{t-1}                  & \mbox{if }t > 1, m_{t-1}^d = 1 \\
\; 0                              & \mbox{if }t = 1
\end{cases}
$$
An example with input matrices is shown in Fig. \ref{fig:archi}. In this example, $S$ is defined as $S = \left\{0, 2, 3, 7, ..., s_T\right\}$.

The goal is to impute the missing values in the incomplete matrix $X$ through the adversarial learning mechanism of GAN with accuracy. The generator tries to make a complete time series matrix $\hat{X}$ by learning the underlying distribution of $X$, and the discriminator tries to match an estimated mask matrix $\hat{M}$ with a mask matrix $M$ by distinguishing whether each element of the 
complete matrix is a real value from $X$ or a fake value from $\hat{X}$. Each element of $\hat{M}$ represents the probability that each element of $\hat{X}$ is a real value. The probability is denoted as $p_t^d \in \left[0, 1\right]$.

\section{Approach}
This section describes how we build an imputation model for multivariate time series data with a focus on the architecture and the overall workflow as shown in Fig. \ref{fig:archi}.

Inspired by existing deep generative neural networks \cite{Goodfellow2014GenerativeAN, yoon2018gain}, STING is composed of two generators (Gs) and one discriminator (D). Two Gs take input time series in the forward and backward directions, each of which generates a time series matrix $\hat{X}$ according to their own designated direction. The generation of both directions complements the lack of information problem in the unidirectional GRU structure suffering from many missing values of time series data. The inside of G is composed of two attention modules and two modified GRU cells \cite{che2018recurrent, luo2018multivariate}. The novel attention modules are able to provide more clues to following GRU modules by calculating the correlation weights of the whole sequence. Then, two GRU cells are designed to deal with the time sequence input with irregular time intervals. On the other hand, D has a relatively simple structure compared to G. It consists of a GRU module for time series processing, a linear layer to reduce the dimension, and finally a sigmoid activation which produces the probability for each element as an output.

In a workflow point of view, our approach starts by making $\bar{X}$ filled with zero to each missing value, then those elements are replaced by $z_t^d$. $M$ serves to inform G of information about which elements of $X$ are observed values or missing values. This allows G to determine which values of the previous timestamp are reliable or not, when repeatedly generating $\hat{x}_t$. $\Delta$ has important information about how much G should refer to the previous hidden states in the case where there are missing values arbitrarily. Based on these four input matrices, G generates an imputed matrix $\hat{X}$ and refines it by filling in the observed values of $X$, as we do not need to generate $x_t^d$ we already know. That is, $\hat{x}_t^d$ is replaced by $x_t^d$ if the element is an observed value. It is the result of imputation and is forwarded to the discriminator.

Meanwhile, D takes $\hat{X}$ as an input and learns to distinguish whether each element is an observed value or not. In this process, the hint matrix $H$ is also provided as an additional input, which informs D of certain parts of $M$ to enforce its attention on the particular components. $H$ reveals some parts of $M$ with $h_t^d=0$ (as a missing value) or $h_t^d=1$ (as an observed value). In addition, $h_t^d=0.5$ implies nothing about $m_t^d$ where D's learning could be concentrated. This is because D has to choose between 0 and 1 for the sample points with 0.5, which comparatively become more difficult learning points and D will focus on getting a better fit. If we do not provide enough information about $M$ to D, several distributions that G could reproduce would all be optimal with respect to D. In other words, we cannot guarantee that G learns the desired distribution uniquely defined by the original data without a hint mechanism. \cite{yoon2018gain} have detailed proofs and theoretical analysis for the hint mechanism. In practice, we could control the amount of hint information about $M$ by varying $H$ in the hint generator. The more hints we provide, the easier D is to learn. We could use this principle to control the learning pace between G and D. Finally, the output of D is $\hat{M}$ where each element represents the probability $p_t^d$ that each element of $\hat{X}$ is an observed value.

By jointly training G and D via a min-max game, G is able to learn the underlying distribution of the original data $X$, and impute the missing values appropriately so that D cannot distinguish them. The ideal result for G is, each $p_t^d$ of $\hat{M}$ corresponding to the generated fake value $\hat{x}_t^d$ is set to 1. In contrast, D aims to accurately distinguish between the real value $x_t^d$ and the fake value $\hat{x}_t^d$. So, the ideal result for D is that each $p_t^d$ corresponding to the fake value $\hat{x}_t^d$ is set to 0, otherwise 1. This implies that D accurately predicts the resulting $\hat{M}$ equal to the mask matrix $M$. As one of the characteristics of STING, D attempts to distinguish which elements of the matrix are real (observed) or fake (imputed), not the entire input matrix itself. This strategy allows D to focus more on the specific classification problem, thereby improving performance.

\subsection{Generator}
STING exploits two types of generators (i.e., forward G and backward G) to account for dependencies in both directions in time as shown in Fig. \ref{fig:archi}. The roles of the Gs are the same except for some parts described in "Consistency loss" (\ref{eq:7}) below. Therefore, only the details of the forward G are described in most of this section, and the description of the backward G is omitted due to the size limitation of the paper. Each G consists of an advanced GRU structure with two types of attention layers (i.e., self-attention and temporal attention), temporal decay layer, and double GRU-cell. Detailed composition and learning objectives are described below.

\textbf{Attention} aims to learn structural dependencies between different coordinates of the input data by finding the most relevant parts of the inputs given a query value and generating a query-specific representation of the inputs. The attention mechanism has been shown to be effective in various fields \cite{bahdanau2014neural, xu2015show, you2016image, zhang2019self} by allowing structural properties of the underlying data distribution to be learned. In a machine translation task, for example, this mechanism is used to measure how much attention should be paid to each word of the sequence of the encoder during decoding. Among various attention algorithms, the scaled dot-product attention \cite{vaswani2017attention} is defined as:
\begin{equation}\label{eq:1}
    \mbox{Attention}(Q, K, V) = softmax(\frac{QK^T}{\sqrt{d_k}})V
\end{equation}
where Q represents the queries, K is the keys, and V is the values. The scale factor $\sqrt{d_k}$ is to avoid overly large values of the inner product, especially when the dimensionality is high. So, the attention function is realized by calculating the weights of correlation called attention scores, between the whole sequence of the encoder as keys and values, and the specific time step of the decoder as a query. In particular, a self-attention module calculates the attention scores between different positions of its own sequence (i.e., Q=K=V) in order to compute representations of the same sequence.

To allow the model to jointly attend to information from different representation sub-spaces at different positions, we further adopt the multi-head attention \cite{vaswani2017attention} defined as:
\begin{equation}\label{eq:2}
    \begin{aligned}
        \mbox{MultiHead}(Q, K, V) = (\mbox{head}_1 \oplus \cdots \oplus \mbox{head}_h)W^O\\
        where\; \mbox{head}_i = \mbox{Attention}(QW_i^Q, KW_i^K, VW_i^V)\\
    \end{aligned}
\end{equation}
where $W_i^Q \in \mathbb{R}^{d_{model} \times d_q}$, $W_i^K \in \mathbb{R}^{d_{model} \times d_k}$, $W_i^V \in \mathbb{R}^{d_{model} \times d_v}$, $W^O \in \mathbb{R}^{hd_v \times d_{model}}$ are the trainable parameter matrices. $\oplus$ denotes the concatenation operation. In our model, $d_{model}$ corresponds to the input dimension of the data, and four parallel attention heads are calculated in the reduced dimension and then concatenated to the original dimension.

\begin{figure}[t]
    \centering
    \includegraphics[width=0.45\textwidth,keepaspectratio]{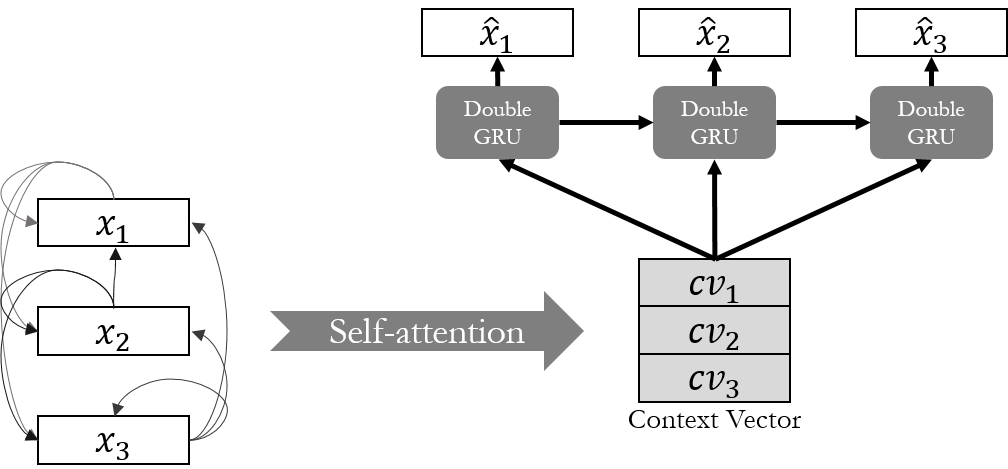}
    \caption{The self-attention mechanism in the imputation process. A context vector is expected to have a good summary of the meaning of the whole input time series (e.g., $x_1$, $x_2$, and $x_3$)}
    \label{fig:self_attention}
\end{figure}

In our work, we adopt the attention mechanism to provide more targeted clues to the imputation problem for two reasons. First, it allows the model to pay attention to time steps with a relatively similar pattern, not just a temporal order. It is suitable especially for time series with a periodicity. A lot of the time series could have a periodic characteristic in a sequence (e.g., weather forecasting, traffic prediction, etc.). Second, it can obtain sufficient information from the entire sequence, not only a few time steps back and forth. If the sequence has many missing values, the quality of the information provided from the whole sequence is improved than only from some time steps. These characteristics are hard to catch in previous works dependent on the temporal order (i.e., RNN methods). So as to take these advantages in our model architecture, the self-attention module is used to convert an input sequence to context vectors (i.e., Q and K are derived from the same input sequence), which is considering the quantitative correlation of the whole sequence as shown in Fig. \ref{fig:self_attention}. Thereafter, the temporal attention module is designed to reflect the correlation between hidden states of a GRU and the context vectors (i.e., Q is hidden states, K is context vectors). These two modules are shown to be effective in our experiments, which will be covered in Section \ref{sec5:rq3}.

\textbf{Temporal decay layer} is to control the influence of the past observations for irregular time intervals in a GRU, which is motivated by \cite{che2018recurrent, luo2018multivariate}. The decay rates should be learned from the data because they differ from variable to variable based on the underlying properties associated with the variables as the missing patterns are unknown and could be complex. That is, the vector of decay rates $\gamma_t$ at $t$ is defined as below.
\begin{equation}\label{eq:3}
    \gamma_t = \exp{\left\{-\max{\left(0,\;W_{\gamma}\delta_t + b_{\gamma}\right)}\right\}}
\end{equation}
where $W_\gamma$ and $b_\gamma$ are trainable parameters, and an exponential negative rectifier is used to keep each decay rate monotonically decreasing in a range between 0 and 1. The function can be replaced with others if only these conditions are met. Intuitively, this has the effect of decaying the extracted features from hidden states rather than raw input variables directly. By learning how much of the previous information that hidden states have will be decayed and utilized, the decay rates adjust the previous hidden states ($h_{t-1}$) to $h_{t-1}^{\prime}$ by element-wise multiplication before entering a GRU-cell.
\begin{equation}\label{eq:4}
    h_{t-1}^{\prime} = \gamma_t \odot h_{t-1}
\end{equation}
where $\odot$ denotes element-wise multiplication.

\textbf{Double GRU} refers to two GRU-cells that are not completely independent but have different purposes. One as a main-cell is responsible for organizing and transferring information to the next, and the other as a generation-cell generates time series vectors $\hat{x}_t$. The generation-cell has its own hidden states (${hg}_t$), which do not pass through a temporal decay layer as described in (\ref{eq:4}). Both of them have a simple GRU structure \cite{cho2014learning}. The main-cell with a temporal decay layer is defined as follows.
\begin{equation}\label{eq:5}
    \begin{aligned}
        r_t &= \sigma\left(W_r x_t + U_r h_{t-1}^{\prime} + b_r\right)\\
        z_t &= \sigma\left(W_z x_t + U_z h_{t-1}^{\prime} + b_z\right)\\
        n_t &= \tanh\left(W_n x_t + U_n\left(r_t \odot h_{t-1}^{\prime}\right) + b_n\right)\\
        h_t &= \left(1-z_t\right) \odot n_t + z_t \odot h_{t-1}^{\prime}\\
    \end{aligned}
\end{equation}
where $x_t$ is the input at $t$, and $r_t$, $z_t$, $n_t$ are the reset, update, and new gates, respectively. Matrices $W_r$, $U_r$, $W_z$, $U_z$, $W_n$, $U_n$ and vectors $b_r$, $b_z$, $b_n$ are the learnable parameters. $\sigma$ is the sigmoid activation, and $\odot$ is the element-wise multiplication. In the case of the generation-cell, $h_{t-1}^{\prime}$, $h_t$ are replaced by ${hg}_{t-1}$, ${hg}_t$, respectively as described in (\ref{eq:5}).

\textbf{Reconstruction loss} is a measure of how well the observed values are reconstructed through G by considering the differences of the observed values in $x_t^d$ and $\hat{x}_t^d$ before those are refined. This loss helps guide the generated data closer to the real distribution of the ground-truth data by inducing G to keep the observed values used as conditions. When obtaining a result of a completely imputed matrix $\hat{X}$, we measure the reconstruction loss ($\mathcal{L}_R$) between $x_t^d$ and $\hat{x}_t^d$ for the observed values. To be specific, if $x_t^d$ is an observed value (i.e., $m_t^d$=1), we directly use it to get the loss and also replace it with the original $x_t^d$ when the output $\hat{x}_t^d$ passes to the next step as an input for the GRU iteratively. In contrast, if $x_t^d$ is a missing value (i.e., $m_t^d$=0), $\mathcal{L}_R$ could not be obtained immediately because the comparison with missing values is impossible. However, STING considers missing values as variables, not constants. So, we replace the missing values with the imputed values to validate those by the future observations. If there is an observed value anywhere in the next of the same sequence, we can get a delayed gradient \cite{cao2018brits}. We measure $\mathcal{L}_R$ in L2-distance as follows.
\begin{equation}\label{eq:6}
    \mathcal{L}_R = \mathbb{E}\left[m_t^d{\left(x_t^d - \hat{x}_t^d\right)}^2\right]
\end{equation}
where for all $t$ and $d$, $m_t^d \in M$, $x_t^d \in X$, $\hat{x}_t^d \in \hat{X}$. It is to note that $\mathcal{L}_R$ should be computed before elements of $\hat{X}$ are refined to $x_t^d$ for the observed values.

\textbf{Consistency loss} can be derived to enforce the prediction in each step to be consistent even in different directions, by means of the structure in which STING consists of two Gs in the forward and backward directions. In addition, consistency loss ($\mathcal{L}_C$) helps the two Gs in the opposite directions interact with each other, so it enhances the learning effect and improves stability by leading to a single result from two matrices. This way also allows obtaining the less delayed gradient of the two directions as described in (\ref{eq:6}) due to less delayed comparison with the closer observation from the forward or backward \cite{cao2018brits}. To measure consistency loss, unlike the forward G, the input matrices are flipped for the backward G during the generating process, and then the output of the backward G (i.e., $\hat{X}_{backward}$) is flipped back to its original sequence for comparison with the output of the forward G (i.e., $\hat{X}_{forward}$). From this, we define $\mathcal{L}_C$ in L1-distance as follows.
\begin{equation}\label{eq:7}
    \mathcal{L}_C = \mathbb{E}\left[\left\vert \hat{x}_{t,forward}^d - \hat{x}_{t,backward}^d \right\vert \right]
\end{equation}
where for all $t$ and $d$, $\hat{x}_{t,forward}^d \in \hat{X}_{forward}$, $\hat{x}_{t,backward}^d \in \hat{X}_{backward}$, and $\hat{X}_{forward}$, $\hat{X}_{backward}$ are imputed data generated from the forward G and backward G, respectively. $\mathcal{L}_C$ should also be computed before elements of $\hat{X}$ are refined for the observed values.

\textbf{Wasserstein loss} proposed in Wasserstein GAN (WGAN) is adopted, which makes a model easier to train than the other GANs by improving the stability of the learning and getting out of the problem of mode collapse \cite{arjovsky2017wasserstein}. However, our problem settings are quite different from the original WGAN because the final results of D are elements of a matrix, not scalar values. Therefore, we derive $\mathcal{L}_{W}$ for the imputation as follows.
\begin{equation}\label{eq:8}
\begin{aligned}
    &\hat{X} = G(Z \mid \bar{X},M,\Delta)\\
    &\hat{M} = D(\hat{X} \mid H)\\
    &\mathcal{L}_{W} = -\mathbb{E}\left[\hat{m}_t^d \mid m_t^d=0 \right]\\
\end{aligned}
\end{equation}
where $G$ and $D$ are the generator and discriminator, respectively. $\odot$ is element-wise multiplication. For all $t$ and $d$, $\hat{m}_t^d \in \hat{M}$, $m_t^d \in M$. $\mathcal{L}_{W}$ is defined only in the case that the input elements are originally missing values.

\begin{figure}[t]
    \centering
    \includegraphics[width=0.45\textwidth,keepaspectratio]{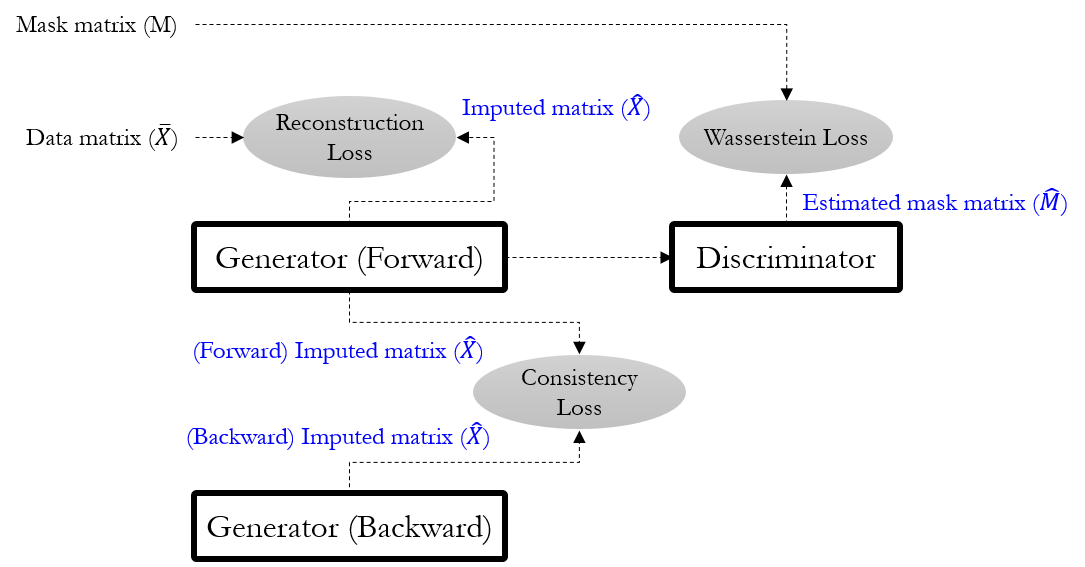}
    \caption{Three types of loss functions. As in Fig. \ref{fig:archi}, the losses calculated by the backward generator are omitted.}
    \label{fig:loss}
\end{figure}

By integrating three losses defined from (\ref{eq:6}) to (\ref{eq:8}) and shown in Fig. \ref{fig:loss}, we can define the total loss of the generators ($\mathcal{L}_G$). In practice, $\mathcal{L}_{W}$ and $\mathcal{L}_R$ are two for each the forward and backward G, and $\mathcal{L}_C$ is one, but only one is described for convenience so that $\mathcal{L}_G$ is defined as.
\begin{equation}\label{eq:9}
    \mathcal{L}_G = \lambda_r\mathcal{L}_R + \lambda_c\mathcal{L}_C + \mathcal{L}_{W}
\end{equation}
where $\lambda_r$ and $\lambda_c$ are hyper-parameters to coordinate learning according to reconstruction loss, consistency loss, respectively.

\subsection{Discriminator}
We introduce a discriminator (D) as an adversary to train G as in the GAN framework. D consists of a GRU module for time series processing, a fully connected layer to reduce the dimension, and finally, a sigmoid activation which produces the probability for each element as an output. We make the structure of D simple compared to G for stable adversarial learning as we experimentally found that D is relatively easy to converge.

The loss function for training D is also relatively simple. For the same reason as G, the loss of conventional Wasserstein GAN could not be applied directly. D's input is a matrix $\hat{X}$ imputed by G which is a combination of imputed values and observed values as elements. In other words, each of these corresponds to a fake value or a real value. This means the loss function should consider two kinds of inputs in a matrix at the same time. For this, we define D's loss function ($\mathcal{L}_D$) as follows.
\begin{equation}\label{eq:10}
    \mathcal{L}_{D} = \mathbb{E}\left[\hat{m}_t^d \mid m_t^d=0 \right] - \mathbb{E}\left[\hat{m}_t^d \mid m_t^d=1 \right]
\end{equation}
where the first term indicates the loss when D takes fake values and the second term is the case of real values as inputs, respectively. When D takes a fake value, it should output 0, or takes a real value, it should output 1. Since for the forward G and backward G, each produces a matrix as an output, D also has two losses corresponding to both directions. But only one is presented for convenience.

\subsection{Searching an Optimal Noise}
Most of the research using GAN \cite{Goodfellow2014GenerativeAN} aims to generate multiple realistic samples by varying a random vector $z$ called "noise". But this may not be applicable to the imputation problem. The imputation method should do two tasks well, that is, not only fill in missing values but also match observed values with accuracy. Because of this specificity, $z$ searching plays an important role. To be specific, the random noise vector $z$ is randomly sampled from latent space such as the Gaussian distribution. This implies that as the input random noise $z$ changes, the generated sample $G(z)$ can change a lot. Even though the generated sample follows the true distribution of the original data, the degree of similarity between $x$ and $G(z)$ could not be large enough. In other words, both distributions could be similar in a broader sense, but individual samples may be different a lot in a specific view. To address this problem and increase the similarity further, it is introduced to find the best matched optimal noise $z^\prime$. Since we already know some conditions (i.e., observed values in a sample), more appropriate $z$ can be found iteratively using those. This method is widely applied in the fields of texture transform and in-painting of image data \cite{gatys2015texture, gatys2016image, yeh2017semantic}, and tabular data imputation \cite{luo2018multivariate}.

Inspired by these works, in our model, we search the optimal noise $z^\prime$ by training $z$ in the inference phase, which generates missing values more suitable for the samples on the distribution of the original data. The learning in the inference phase does not train parameters of a model but performs back-propagation to input $z_t$ updated every iteration. Learning for $z_t$ uses the same loss as training G ($\mathcal{L}_G$) in (\ref{eq:9}). That is, gradient $-\frac{\partial \mathcal{L}_G}{\partial z_t}$ repeatedly updates $z_t$ to search a more suitable one. After searching the optimal noise $z_t^\prime$ as one of the inputs, two imputed matrices are generated by the forward and backward G, respectively. Finally, we use the mean of two matrices to determine the final result of imputation.

\section{Experiments} \label{sec:experiments}
In this section, we conduct experiments with the purpose to prove the efficacy of our proposed STING model by answering the following research questions:
\begin{itemize}
\item \textbf{RQ1} Does the STING outperform other state-of-the-art imputation methods?
\item \textbf{RQ2} How does the STING work for the downstream task as a post-imputation?
\item \textbf{RQ3} Which module is most influential in improving performance in STING?
\end{itemize}

In the following, we first describe the datasets and baseline methods used in the experiments. Then, we compare the proposed STING with other comparative methods and make a detailed analysis of STING under two different experimental settings. Finally, we conduct an ablation study to analyze the impact of the main modules in STING.

Regarding the details of the basic setting of the experiments, min-max normalization was applied during the experiments for all datasets. All experiments were repeated 10 times and the mean of accuracy was reported to account for any kind of randomness during the experiments. The generators were pre-trained for 10 epochs with $\mathcal{L}_R$ and $\mathcal{L}_C$ in (\ref{eq:6}) and (\ref{eq:7}). We experimentally found that if the generators are trained a little in advance, they could converge faster and get better performance. Then, the generators and the discriminator were updated one by one alternately at every iteration. We trained our model with an Adam optimizer. The learning rates of the generators and the discriminator were 0.001 and 0.0001, respectively. The hint ratio given to the discriminator was fixed at 0.1. The batch size was set to 128. As the hyper-parameters for loss in (\ref{eq:9}), $\lambda_r$ and $\lambda_c$ were set to 10 and 1. We implemented our model in PyTorch and performed all training on a single 2080Ti GPU with 11GB of RAM.

\subsection{Datasets}

\textbf{PhysioNet Challenge 2012 Dataset (PhysioNet)} - comes from PhysioNet Challenge 2012\footnote{https://www.physionet.org/content/challenge-2012} \cite{silva2012predicting, goldberger2000components} which aims to develop methods for patient-specific prediction of in-hospital mortality. It consists of records from 12,000 multivariate clinical time series from intensive care unit (ICU) stays. We use training set A (4,000 ICU stays) of the whole dataset. The preprocessed dataset for the experiments has a total of 192,000 samples. Each sample contains 37 variables such as DiasABP, HR, Na, Lactate, etc. over 48 hours. There are 554 (13.85\%) patients with a positive mortality label. As this dataset has a high missing rate (80.53\%) and is very sparse, it is difficult to simply perform downstream tasks such as the mortality prediction. So, many previous works have experimented on this dataset to evaluate the performance of imputation or post-imputation tasks.

\textbf{KDD CUP Challenge 2018 Dataset (Air Quality)}\footnote{https://archive.ics.uci.edu/ml/datasets/Beijing+Multi-Site+Air-Quality+Data} - accessible from UCI Machine Learning Repository \cite{Dua:2019}, is a public air quality dataset and used in KDD CUP Challenge 2018 \cite{zhang2017cautionary} to accurately forecast air quality indices (AQIs) of the future 48 hours. The records have a total of 12 variables such as PM2.5, PM10, SO2, etc. from 12 monitoring stations in Beijing. The time period is from March 1, 2013 to February 28, 2017 and the variables were measured every hour. The total number of samples is 420,768, and there are some missing values (1.43\%).

\textbf{Gas Sensor Array Temperature Modulation Dataset (Gas Sensor)}\footnote{https://archive.ics.uci.edu/ml/datasets/Gas+sensor-array+temperature+modulation} - accessible from UCI Machine Learning Repository, contains 14 temperature-modulated metal oxide (MOX) gas sensors exposed to dynamic mixtures of carbon monoxide (CO) and humid synthetic air in a gas chamber for 3 weeks \cite{burgues2018estimation, burgues2018multivariate}. We use one day of the whole dataset for the experiments. The number of samples is 295,704. Each sample consists of 20 variables including CO concentration inside the gas chamber. Unlike the other datasets, all samples are completely observed.

\begin{figure*}[t]
    \centering

    \begin{subfigure}{0.49\textwidth}
    \centering
    \includegraphics[width=\textwidth,keepaspectratio]{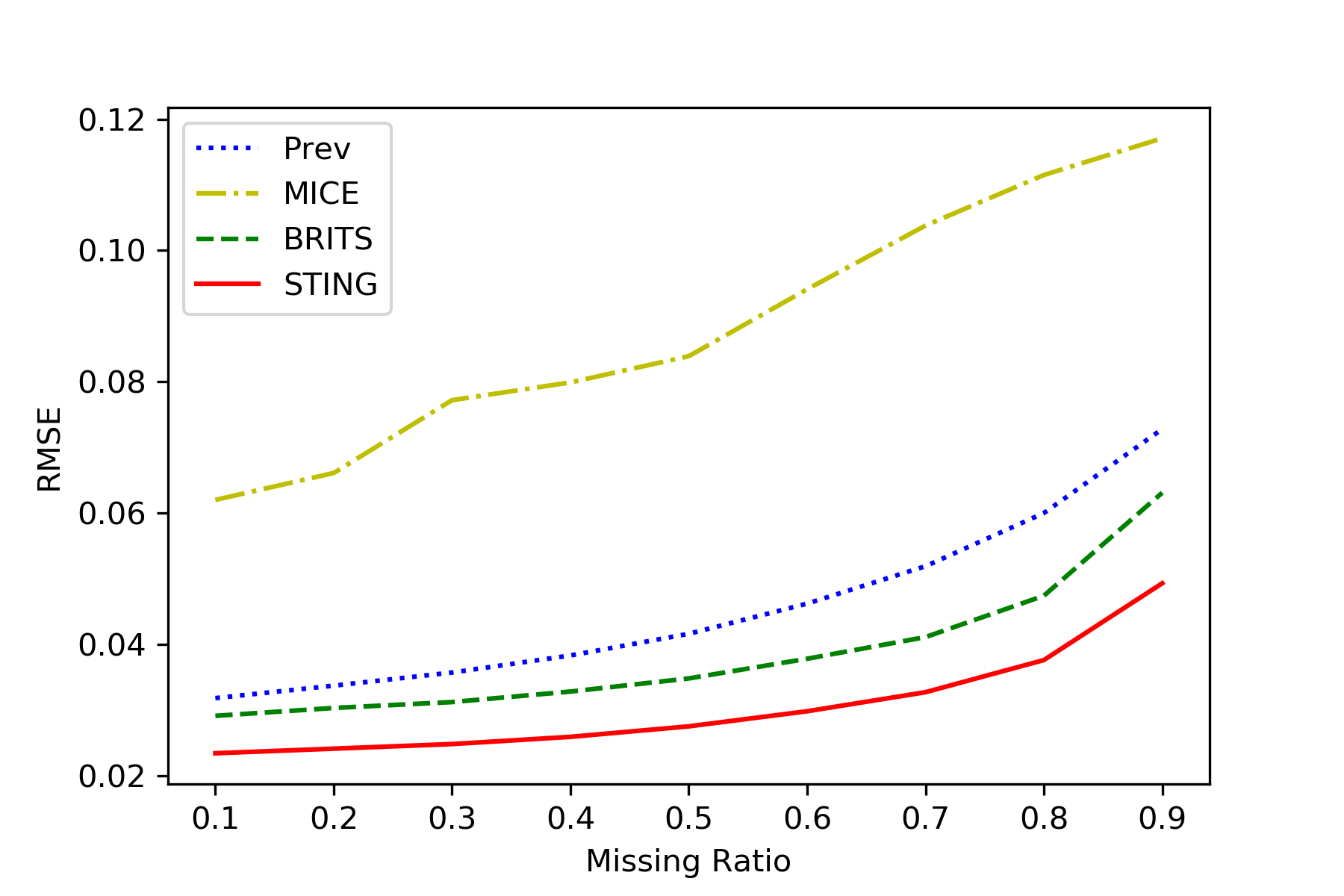}
    \subcaption{Air Quality}
    \end{subfigure}
    \begin{subfigure}{0.49\textwidth}
    \centering
    \includegraphics[width=\textwidth,keepaspectratio]{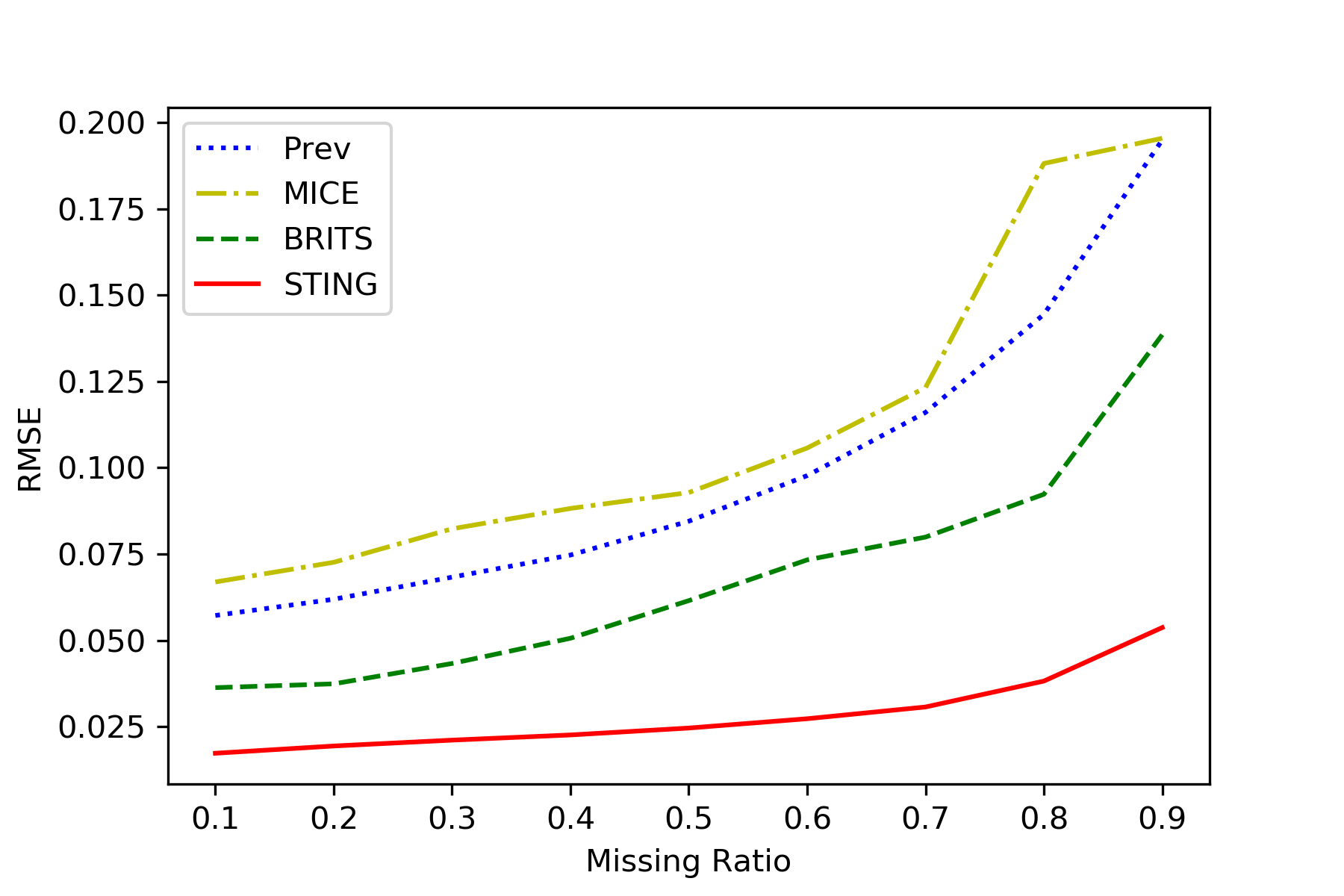}
    \subcaption{Gas Sensor}
    \end{subfigure}

    \caption{Direct imputation performances on two datasets while varying the ground-truth ratio (since these datasets are almost complete, the ground-truth ratio and the missing ratio are almost the same). As comparative methods, three representative models with the best performance except STING were selected from each type of baselines.}
    \label{fig:direct_graph}
\end{figure*}

\subsection{Baselines}
In order to evaluate the performance of our model, we compare it with the following representative baselines. There are three types of models: statistics-based (Stats-based), machine learning-based (ML-based), and neural network-based (NN-based) models. We implemented ML-based models based on the python package $sklearn$ and $fancyimpute$. The experimental settings such as hyperparameters of the NN-based models were set according to the corresponding papers, respectively.

\begin{itemize}
\item \textbf{Mean} simply fills the missing values with the global means of the corresponding variables.
\item \textbf{Previous Value Filling (Prev)} fills the missing values with the previously observed values. This method can be very simple and computationally efficient for imputation because of the characteristic of time series.
\item \textbf{KNN} \cite{troyanskaya2001missing} uses K-Nearest Neighbors which finds the similar 10 samples, then imputes the missing values with means of these samples.
\item \textbf{Matrix Factorization (MF)} \cite{koren2009matrix} directly factorizes the incomplete matrix into two low-rank matrices solved by gradient descent, then imputes the missing values by matrix completion.
\item \textbf{Multiple Imputation by Chained Equations (MICE)} \cite{azur2011multiple, buuren2010mice} models each feature with missing values as a function of other features iteratively, and uses that estimate for imputation.
\item \textbf{Generative Adversarial Imputation Nets (GAIN)} \cite{yoon2018gain} imputes the missing values conditioned on what is actually observed using GAN.
\item \textbf{Gated Recurrent Unit with Decay (GRU-D)} \cite{che2018recurrent} based on GRU, imputes the missing values by using a decay mechanism for irregular time intervals.
\item \textbf{End-to-End Generative Adversarial Network (E$^2$GAN)} \cite{luo2019e2gan} is based on GAN and the generator has an auto-encoder structure. That is, it can optimize the low-dimensional vector while learning the distribution of the original time series.
\item \textbf{Bidirectional Recurrent Imputation for Time Series (BRITS)} \cite{cao2018brits} adapts bidirectional RNN for imputing missing values without any specific assumption over the dataset.
\end{itemize}

\begin{table}[t]
\centering
\caption{Overall model performances in terms of RMSE (the smaller, the better). The results of the best performing baseline and the best performer in each column are underlined and boldfaced, respectively. The ground-truth ratio is set to 20\% of the observed values. The RMSE is measured after min-max normalization.}
\label{tb:1}
\begin{tabular}{ccccc}
\toprule
\multicolumn{2}{c}{} & PhysioNet      & Air Quality     & Gas Sensor    \\
\toprule
\multirow{2}{*}{Stats-based}
& Mean     & 0.1037 & 0.1203 & 0.2000 \\
& Prev     & \underline{0.0825} & \underline{0.0336} & \underline{0.0619} \\
\midrule
\multirow{3}{*}{ML-based}
& KNN      & 0.1137 & 0.1157 & \underline{0.0379} \\
& MF       & 0.1033 & 0.1008 & 0.1355 \\
& MICE     & \underline{0.0986} & \underline{0.0659} & 0.0726 \\
\midrule
\multirow{5}{*}{NN-based}
& GAIN     & 0.1300 & 0.1148 & 0.0722 \\
& GRU-D    & 0.0830 & 0.0377 & 0.0630 \\
& E$^2$GAN & 0.0638 & 0.0379 & \underline{0.0374} \\
& BRITS    & \underline{0.0553} & \underline{0.0301} & \underline{0.0374} \\
& STING    & \textbf{0.0531} & \textbf{0.0238} & \textbf{0.0153} \\
\bottomrule
\end{tabular}
\end{table}

\subsection{Direct Evaluation of Imputation Performance (RQ1)}
In this experiment, we aim to directly evaluate the imputation performance with baselines and STING. We randomly eliminated a certain ratio of the observed values to be used as the ground-truth. In consequence, the remaining observed values were used as the training data for the imputation models. Once the imputation learning of each model was done, we inferred complete data by imputing missing values via each model. After that, by comparing imputed values with the ground-truth, the performance of each model was measured by Root Mean Squared Error (RMSE) indicating that the lower the value, the better the performance. In the first experiment, we set a ratio of the ground-truth to 20\% and compared the performances of all the models on three datasets. In the second experiment, we evaluated the performances of the best performing baselines by varying the ground-truth ratio from 10\% to 90\%.

Table \ref{tb:1} summarizes the results of the first experiment. We can see that STING achieves the best performance on all datasets. In comparison to the best performing baseline (i.e., BRITS), error improvement rates of the results are 4.0\%, 21.0\%, and 59.1\%, respectively. Prev as one of the statistical methods has a fairly high performance. This indicates that simply imputing with the previous values could have better performance than the other models which take a long time and are complicated. It is reasonable because the dependence on the previous samples is high due to the nature of the time series data collection. In particular, for a dataset with little change over time, filling with previous values can have a great effect with little cost. In addition, Prev shows a similar tendency of the performance to GRU-D. This is a quite logical result because GRU-D learns the ratio between the mean and the previous value, then imputes the missing values. Among the ML-based models, KNN shows the best performance similar to BRITS on the Gas Sensor dataset, but the performance is not stable on other datasets. On the other hand, MICE stably shows relatively good performance on all datasets. GAIN motivated by GAN does not perform well on the time series datasets as there is no suitable learning strategy for time series.

Fig. \ref{fig:direct_graph} shows the performance results of the second experiment when the missing rate of the two datasets is varied. It is to note that we exclude PhysioNet for this experiment because it has a very high missing rate (80.53\%). As the missing ratio increases, the training data used for imputation learning decreases on the contrary, so the performances of all models decrease. Nevertheless, STING achieves the best performance in all conditions, showing a slow decline in performance. That is, the more the dataset contains missing values, the less sensitive STING imputes them compared to others. From the results, we could confirm STING takes advantage of the attention mechanism because it can utilize a relatively large amount of information by referring to the whole sequence.

\subsection{Indirect Evaluation of Imputation Performance (RQ2)}
In this experiment, we aim to indirectly evaluate imputation performance by the results of solving downstream tasks. If the distributions of original data and imputed data are similar, their results of downstream tasks would be similar. Therefore, we can see how well incomplete data were imputed indirectly through the prediction results. We only exploit the Gas Sensor dataset which is originally complete data for the stable learning of the prediction model. Moreover, this data can assume an ideal imputation model by which all missing values have been imputed with the ground-truth, so that we can get an upper bound prediction performance. The ideal imputation model is also included as a baseline in this experiment. It is to note that we do not train simultaneously imputation and prediction tasks at the same time because our purpose is to measure the imputation performance, not to improve the prediction performance. After imputing the test data except for labels, those are inferred with the prediction model already learned on the complete train data. This is intended to be a fair comparison of models with typical setups where there may be no clear labels or missing values on the labels.

The procedure of this experiment is as follows. For independent settings between the imputation model and regression model, the dataset is initially divided into 80\% training data and 20\% test data. Using this complete training data, we first train a regression model that predicts the CO concentration as a target. As the regression model, we built a simple GRU with two layers and a dropout with 0.3, and finally having a fully connected layer.
Then, the training data is made to have missing values completely at random in a certain ratio. After training our method and baselines on the training data, the resulting models impute the missing values of the test data except for the target so that we can get different imputed test datasets corresponding to each model. Then, the regression model is used to predict the target based on imputed test data. Finally, we measure the RMSE results between the predicted target values and the actual ones.
This evaluation method allows us to determine if the imputed data follow the distribution of the original data under the assumption that similar distributions get similar performances. In other words, we can know which imputed data are closer to the true distribution by comparing how close they are to the regression result of the original test data (i.e., imputed by the ideal model). It is to note that we do not aim to achieve state-of-the-art prediction performance.

\begin{figure}[t]
    \centering
    \includegraphics[width=0.49\textwidth,keepaspectratio]{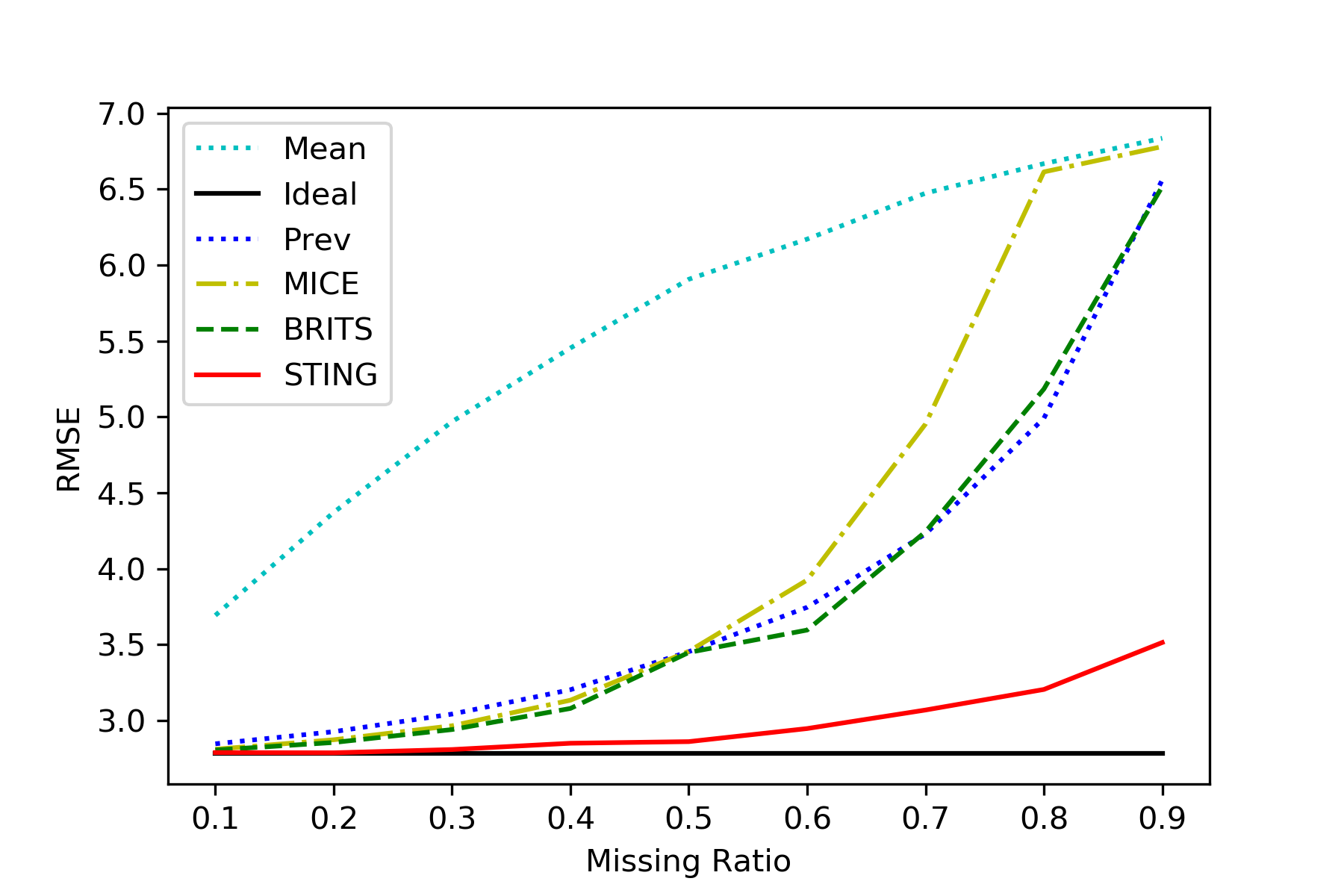}
    \caption{Indirect imputation performance results on Gas Sensor while varying the ground-truth ratio.}
    \label{fig:indirect_graph}
\end{figure}

Fig. \ref{fig:indirect_graph} shows the RMSE results of the regression model while varying the ground-truth ratio on each imputed data. The ideal model is able to impute the original test data regardless of any missing ratio, so it has the lowest and constant RMSE. Among the comparative imputation models, STING achieves the best performance at any ratio. On the other hand, Mean shows the worst performance, showing how inefficient it is to impute the time series with the mean values. Interestingly, there is a clear difference between the missing ratio of around 0.5. Previously, most models maintain similar and good performances below 3.5, but the RMSE increases dramatically afterward. When the ground-truth is 0.9, STING shows a relatively small increase in error despite the poor conditions. There seems to be an advantage to using GANs to generate new time series and to utilize information from the entire sequence due to the attention mechanism.

\begin{table}[t]
\centering
\caption{The ablation study on the attention mechanism, searching an optimal noise $z^\prime$, and the backward generator, respectively. The rate of increase in the error is indicated in parentheses (\%).}
\label{tb:2}
\begin{tabular}{llll}
\toprule
    & PhysioNet & Air Quality & Gas Sensor \\
\toprule
\multicolumn{1}{c}{STING} & \multicolumn{1}{c}{0.0531} & \multicolumn{1}{c}{0.0238} & \multicolumn{1}{c}{0.0153} \\
\midrule
w/o attention & 0.0559 (5\%)  & \textbf{0.0294 (23\%)} & \textbf{0.0355 (132\%)}  \\ 
w/o optimal $z^\prime$   & 0.0553 (4\%)  & 0.0250 (5\%)  & 0.0193 (26\%) \\
w/o backward  & \textbf{0.0583 (10\%)} & 0.0269 (13\%)  & 0.0189 (23\%) \\
\bottomrule
\end{tabular}
\end{table}

\subsection{Ablation Study (RQ3)} \label{sec5:rq3}
The potential sources of efficacy for the STING are: two attention modules, a method for searching an optimal noise $z^\prime$, and a backward generator, respectively. To understand how each key feature affects the performance improvement of STING, we conducted an ablation study and compared the performances of the resulting architecture to the overall STING architecture. So, we experimented with three models by excluding just one function from the STING on three datasets. The ground-truth ratio was set to 20\% of observed values, and the RMSE with imputed values was measured.

Table \ref{tb:2} lists the results of the ablation study, where the RMSE increases in all cases. The result of removing the backward generator shows the highest increase in error rate on PhysioNet. On the other hand, the result removing the attention modules shows the highest increase in error rates on Gas Sensor and Air Quality. This implies that the proposed attention modules perform a relatively important function in STING. That is, the process of learning the correlations of the whole sequence produces significantly important information for imputation in any dataset. On the other hand, the module for searching an optimal noise $z^\prime$ has a relatively minor effect. This indicates that even though $z$ is randomly generated, it is possible to make a sample that matches well from the actual distribution of the original data because the observed values in $X$ are used as a condition to G in STING. In other words, STING without an optimal noise might efficiently learn the distribution of original time series, so this process plays a complementary role.

In summary, we demonstrated the efficacy of STING through comparative experiments with other models. There are two main factors attributable to the remarkable performance even in poor conditions.
First of all, we could confirm that the GAN mechanism works well to generate samples that follow the true distribution of the original time series. We described how STING should be configured to converge to the desired distribution. In particular, we set the problem of discriminator more delicately with Wasserstein distance to maximize the adversarial learning effect, compared to previous works based on GAN.
As the second key factor, we could find that the proposed attention mechanism is effective in the imputation task. 
STING could acquire a large amount of information by paying attention to certain periodic patterns or some time steps to get more information. This effect becomes more evident as the missing rate increases, clearly showing the benefits of retaining large amounts of information during imputation.

\section{Conclusion}
In this paper, we proposed STING, a novel imputation method for multivariate time series data based on generative adversarial networks and bidirectional recurrent neural networks to learn latent representations of time series. We also proposed the novel self-attention and temporal attention mechanism to capture the weighted correlations of the whole sequence and prevent potential bias from unrelated time steps. Various experiments on real-world datasets show that STING outperforms traditional state-of-the-art methods in terms of both imputation accuracy and downstream performance with the imputed values. Future work will investigate the imputation of the more general data including a categorical type. Generating categorical data is a particularly difficult problem for GANs. Inspecting its possibility of generating and imputing categorical data is a topic for future research.


\bibliographystyle{Bibliography/IEEEtran}
\bibliography{Bibliography/refer}

\end{document}